\documentclass[graybox, natbib, nosecnum]{svmult}
\bibpunct{(}{)}{;}{a}{}{,} 

\usepackage{mathptmx}       
\usepackage{helvet}         
\usepackage{courier}        
\usepackage{type1cm}        

\usepackage{graphicx}        
\usepackage{multicol}        
\usepackage[bottom]{footmisc}
\usepackage[normalem]{ulem}	
\usepackage{hyperref}  
\usepackage{soul}   
\usepackage{float}
\usepackage{amsmath}
\usepackage[hyphenbreaks]{breakurl}
\usepackage{xurl}
 
\usepackage{caption}
\usepackage{subcaption}
\usepackage{array}
\usepackage{xcolor}
\usepackage{enumerate}
\usepackage{todonotes}

\begin{document}
\title*{Convolutional Neural Networks Applied to Modification of Images}
\author{Carlos I. Aguirre-Velez \thanks{corresponding author} and Jos\'e Antonio Arciniega-Nev\'arez and Eric Dolores Cuenca}
\institute{First Author \at Instituto Politécnico Nacional CICATA U. Legaria, 
Ciudad de México, México \email{caguirre@ipn.mx}
\and Jos\'e Antonio Arciniega-Nevárez \at División de Ingenier\'ias, Universidad de Guananjuato, Guanajuato, M\'exico \email{ja.arciniega@ugto.mx}
\and Eric Dolores-Cuenca \at Department of mathematics Yonsei University 50 Yonsei-Ro, Seodaemun-Gu, Seoul, Korea
 \email{eric.rubiel@yonsei.ac.kr}}
%
%
\maketitle
\abstract{ The reader will learn how digital images are edited using linear algebra and calculus. Starting from the concept of filter towards machine learning techniques such as convolutional neural networks.
}

\section{Keywords}
Machine Learning, Artificial Intelligence, Convolutional Neural Networks, Diffusion Models, Text To Image,
Please provide 5-10 keywords

\section{Introduction}
Nowadays, it is common to manipulate images by a single click in photo applications for smartphones. Images, videos, and even live streams can be modified using diverse filters without programming knowledge. People do not imagine that all these wonderful digital applications for image enhancement use mathematical algorithms.

In the previous time to the digital era, the images were modified by several ingenious analogue techniques. The roll film reigned in those days when both photographers and cinematographers performed different tricks to edit images and to carried out the special effects of cinema.

In 1964, pictures of the moon transmitted by Ranger 7, a lunar device built to take high-resolution photographs of the moon before impacting the lunar surface, were processed by a computer in order to correct inherent distortion of the image. The lessons learned from these images served as a basis for improving the methods used and for improving and restoring images from other lunar missions~\cite{Picture, Digital}.

The development of digital image processing was possible due to a series of key advances that have led to powerful computers. Some of the most important facts may be summarized as:  
\begin{enumerate}[a)]
\item the invent of transistors, integrated circuits, and microprocessors as the fundamental bricks of hardware building;
\item development of programming code languages and operating systems;
\item the invent of the charge-coupled devices (CCD) and the metal-oxide-semiconductors (CMOS) sensors as basic devices that can capture electromagnetic energy from the visible world and transform the electromagnetic energy into bits~\cite{scientific};
\item the compression algorithms that store a huge amount of information in a smaller file size; 
\item  the introduction of the personal computers for massive use;
\item the development of colorful displays based on LED’s technology and the enhance of its resolution;
\item the development of graphics processing units (GPUs) for work on 3-D applications, and image processing applications involving large-scale matrix implementations.
\end{enumerate}

Digital image processing has many advantages over analog image processing. Mathematical algorithms allow the modification of images for diverse applications such as spatial exploration, agriculture, medical science, industry, commerce, security, etc. In a digital image, a pixel is the smallest controllable element of a display screen. Image enhancement involves modifying {the pixel} values to improve the quality of the image mainly for two applications: improvement information for human interpretation and autonomous machine perception.
The processes are of low level type, middle level type and high level type. Low level is the processing related to reduce noise, contrast enhancement and image sharpening. The Middle level relates to the attributes from the image such as edges, contours, some details and special features. And a high level involves information for computational treatment.

Since the 1980s, the seeds of Artificial Intelligence (AI) began to be laid. But the computing power, both in processing and storage, was insufficient to implement AI at the time. An algorithm of AI needs to process and store a large amount of data, either during training of the algorithm or because the AI algorithm generates the data. Since the new millennium, these two prerequisites have been solved to the point of the necessity of companies dedicated to data collection. AI has become a tool to solve different problems, with success rate based on available information. The mathematical tools needed for the development of AI are taught in the first years of math undergraduate courses, but the algorithms are much more sophisticated than they appear. In this chapter, we shall introduce the mathematical models for data processing that were used before AI and, in a second part, we shall show the general structure of some AI algorithms dedicated to image processing. In particular, this will provide the basis for Convolutional Neural Networks (CNN).

\section{\textit{Mathematical basis of image processing}}
We may say that the main difference between an analog and a digital image is the material and the form in which they are stored. Although the analog image is considered to be a better representation of what is observed, technological advances have made the digital image easier to use, store and process, so that nowadays the analog image has practically disappeared.

The introduction to the mathematical tools to process digital images is given in the following sections.

\subsection{Histograms}
There are different forms of image manipulation, one of the simplest is the histogram which extracts the most basic information from an input image: In fact, a histogram is a graph showing the number of pixels in an image at each different intensity value found in that image. Remember that 256 is the range of gray levels available for displaying an image. Different software let us manipulate histograms manually or by custom functions. Manipulation is generally carried out in basic statistics parameters (mean, variance, median), these operations modify graphic information making an equalization to enhance an image. Some features are easy to adjust in intensity range using the histogram such as contrast, brightness, saturation, and luminance. These kinds of operations are called point operations (see Figure~\ref{fig:orginal-bajo} and Figure~\ref{fig:modified-bajo}).

\begin{minipage}{0.4\textwidth}
\begin{figure}[H]
     \centering
     \begin{subfigure}[b]{0.4\textwidth}
         \centering
         \includegraphics[width=\textwidth]{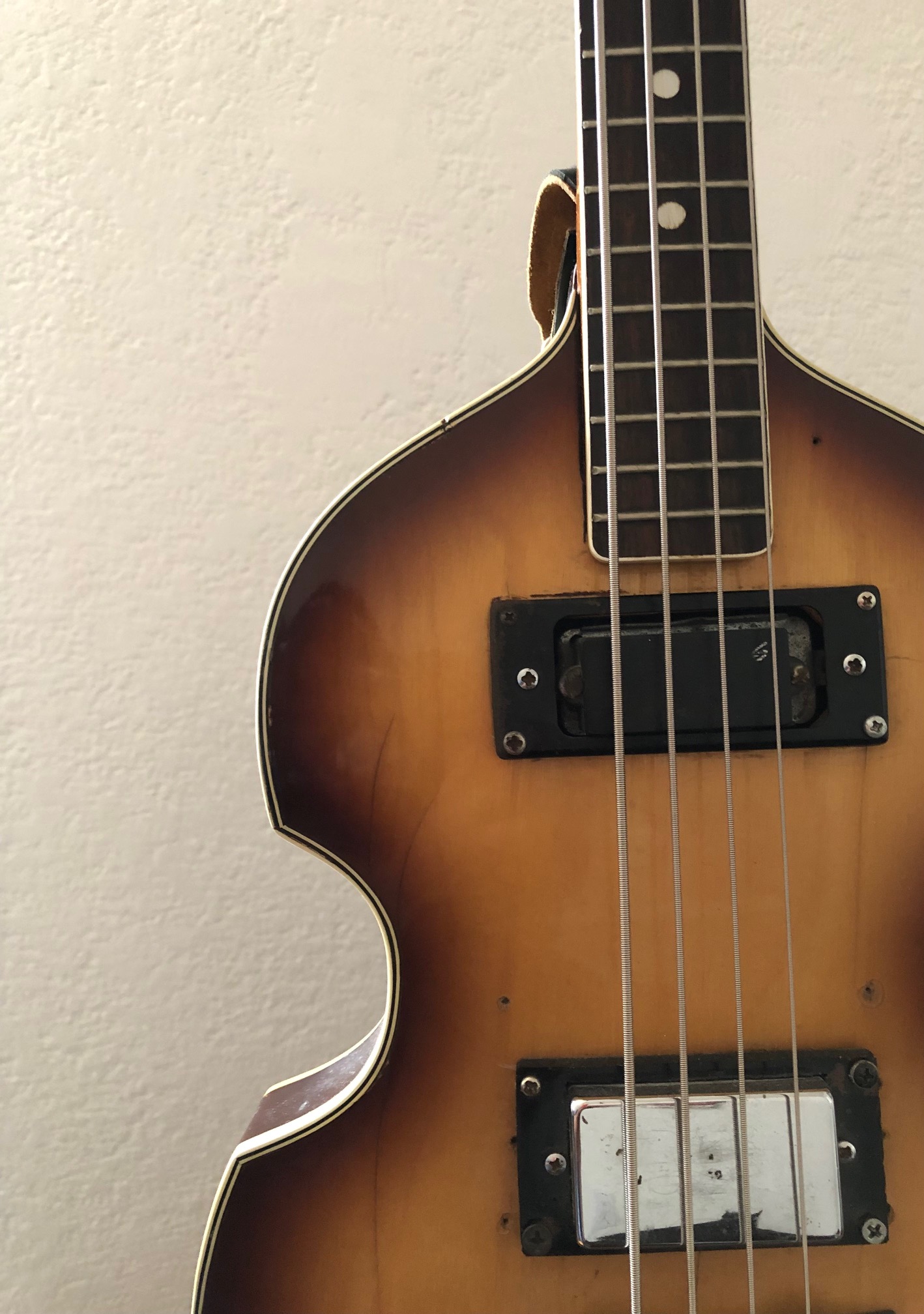}
     \end{subfigure}
     \hfill
     \begin{subfigure}[b]{0.4\textwidth}
         \centering
         \includegraphics[width=\textwidth]{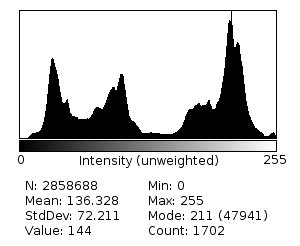}
    \end{subfigure}
     \caption{Original figure and its histogram}
         \label{fig:orginal-bajo}
\end{figure}
\end{minipage}
\hfill
\begin{minipage}{0.4\textwidth}
\begin{figure}[H]
     \centering
     \begin{subfigure}[b]{0.4\textwidth}
         \centering
         \includegraphics[width=\textwidth]{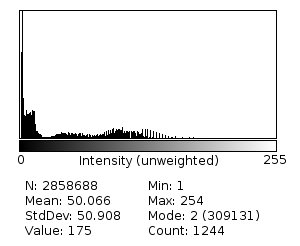}
     \end{subfigure}
     \hfill
     \begin{subfigure}[b]{0.4\textwidth}
         \centering
         \includegraphics[width=\textwidth]{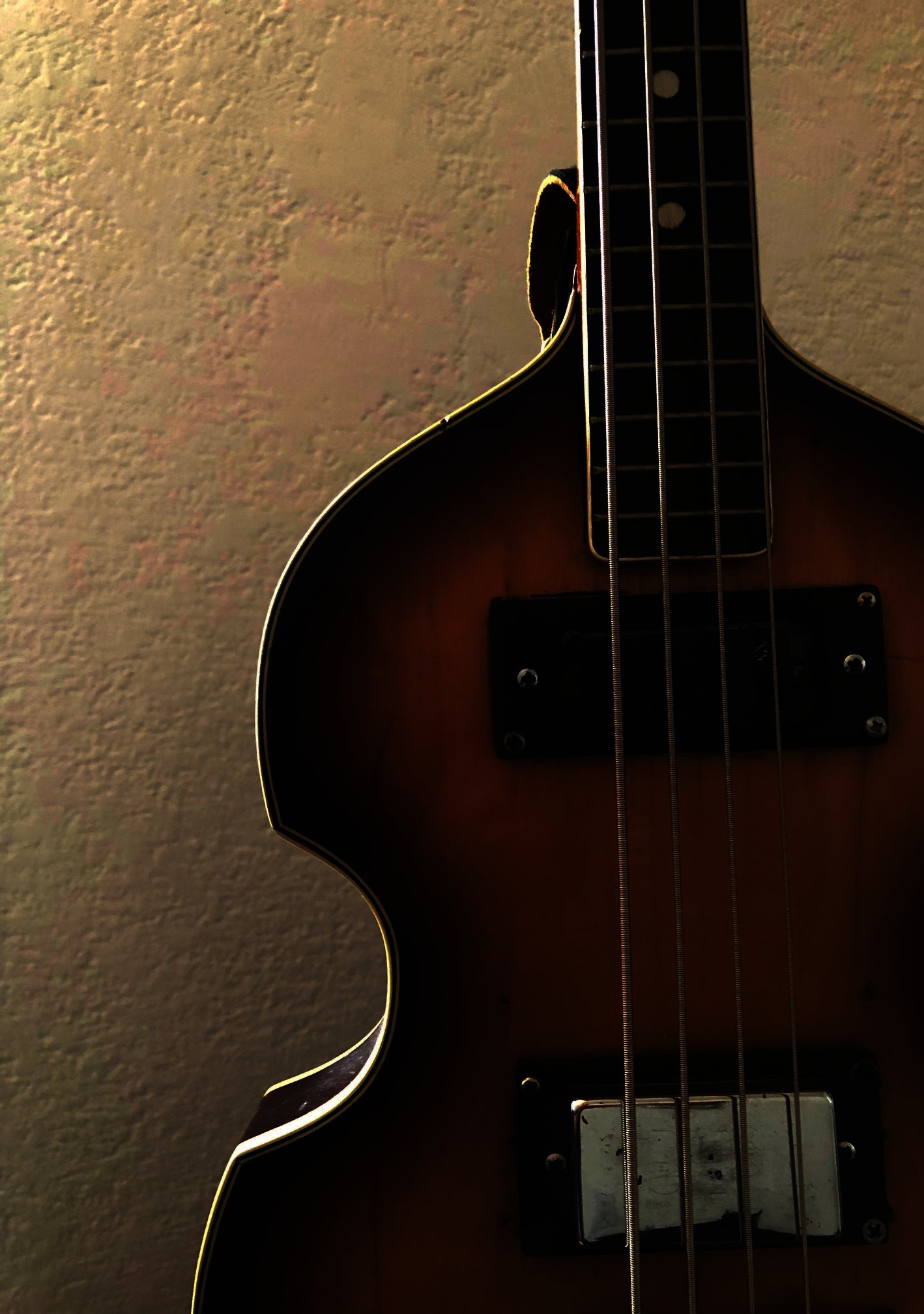}
         
     \end{subfigure}
     \caption{Modified image and its corresponding histogram}
     \label{fig:modified-bajo}
\end{figure}

\end{minipage}

\begin{figure}[H]
\centering
\caption{Images and its histograms. a) original image, b) histogram of original image,
c) modified image, d) histogram of modified image.}    
\end{figure} 
\subsection{Images as matrices}
Another method is to consider an image as a two dimensional function $f(x,y)$ related to the intensity or gray level of each pixel in that coordinates, in other words, an image is a single matrix (a two-dimensional array), the coordinates $(x,y)$ are the \textbf{spatial domain} and $f(x,y)$ is called the \textbf{frequency domain}. Therefore transformations can be applied to these matrices. Applying two transformations is equal to taking the product of the two corresponding matrices operating the image. Because matrix multiplication is associative, multiple transformations can be combined into a single transformation by multiplying the matrix of each individual transformation in the order that the transformations are done. 

It is important to make a clear distinction between elementwise and matrix operations. For example, consider the following two images ($2\times2$ matrices):
\begin{equation}
\begin{pmatrix}a_{11} & a_{12} \\a_{21} & a_{22}\end{pmatrix}, \quad
\begin{pmatrix}b_{11} & b_{12} \\b_{21} & b_{22} \end{pmatrix}
\end{equation}
The elementwise product of these two images is
\begin{equation}\label{matrix-elementwise-prod}
\begin{pmatrix}a_{11} & a_{12} \\a_{21} & a_{22}\end{pmatrix}
\begin{pmatrix}b_{11} & b_{12} \\b_{21} & b_{22} \end{pmatrix}
=
\begin{pmatrix}a_{11}b_{11}& a_{12}b_{12}  \\a_{21}b_{21} & a_{22}b_{22} 
 \end{pmatrix}
\end{equation}

That is, the elementwise product is obtained by multiplying pairs of corresponding pixels. On the other hand, the matrix product (or inner product) of the images is formed using the rules of matrix multiplication:

\begin{equation}\label{matrix-prod}
\begin{pmatrix}a_{11} & a_{12} \\a_{21} & a_{22}\end{pmatrix}
\begin{pmatrix}b_{11} & b_{12} \\b_{21} & b_{22} \end{pmatrix}
=
\begin{pmatrix}a_{11}b_{11}+a_{12}b_{21} & a_{11}b_{12}+a_{12}b_{22}  \\a_{21}b_{11}+a_{22}b_{21} & a_{21}b_{12}+a_{22}b_{22} 
 \end{pmatrix}
\end{equation}

\begin{figure}[H]
     \centering
     \begin{subfigure}[b]{0.3\textwidth}
         \centering
         \includegraphics[width=\textwidth]{img_c/bajo_original.jpg}
         \caption{Original image}
     \end{subfigure}
     \hfill
     \begin{subfigure}[b]{0.3\textwidth}
         \centering
         \includegraphics[width=\textwidth]{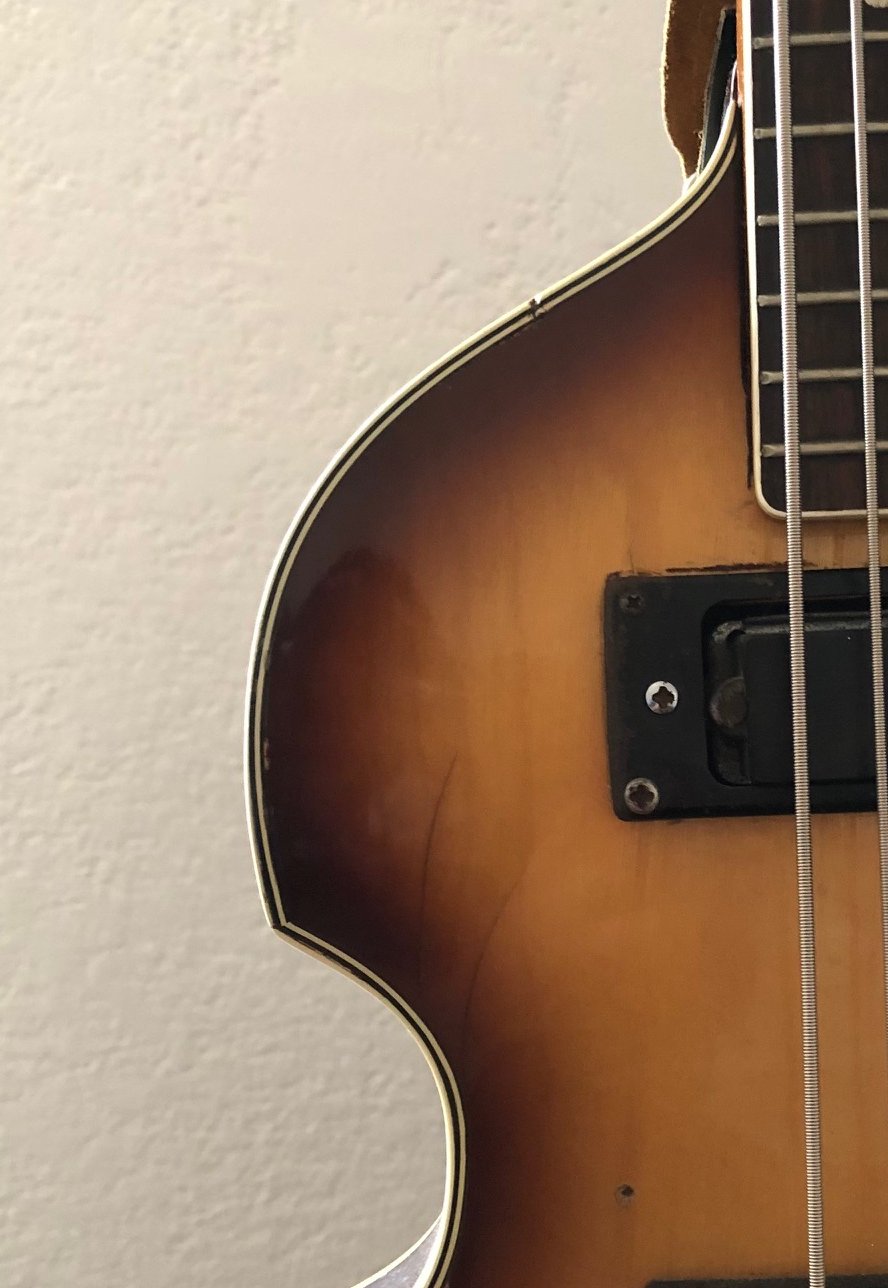}
         \caption{Scaled image}
    \end{subfigure}
    \hfill
    \begin{subfigure}[b]{0.3\textwidth}
         \centering
         \includegraphics[width=\textwidth]{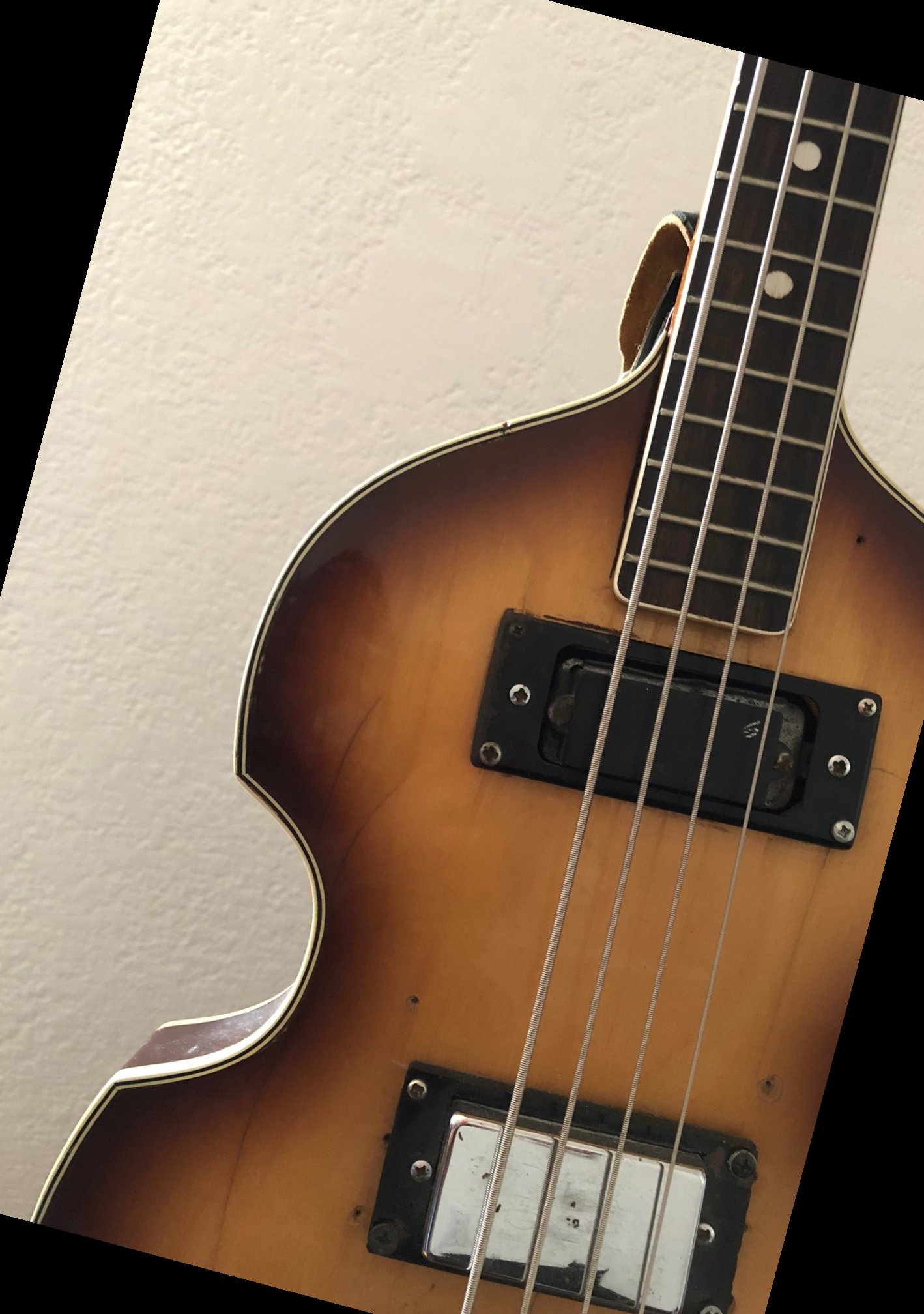}
         \caption{Rotated image}
    \end{subfigure}
     \caption{Operations}
         \label{fig:Operations}
\end{figure}

Some common basic operations on images are the so-called affine transformations: scaling, reflection, shrinking, rotating, translation, and shearing (see Figure~\ref{fig:Operations}).  

These transformations are called \textbf{geometric transformations} and they can be performed by matrix multiplications. We can classify digital images as follows: 1) Spatial transformation of coordinates. 2) Intensity interpolation that assigns intensity values to the spatially transformed pixels. We can denote any pixel by its spacial coordinates $(x,y)$ in the matrix, it also storages the intensity of the pixel. Then a transformation $T$ of coordinates may be expressed as

\begin{equation}\label{trasformation}
\begin{pmatrix}x' \\y'\end{pmatrix}
=
\begin{pmatrix}t_{11} & t_{12} \\t_{21} & t_{22}\end{pmatrix}
\begin{pmatrix}x \\y\end{pmatrix}
\end{equation}

Equation~\eqref{trasformation} can be used to express one of the transformations mentioned before, except translation, which would require that a constant non-zero vector must be added to the right side of the equation. It is possible to use homogeneous coordinates to express all four affine transformations using a single $3\times 3$ matrix, but we do not explain this here.

This accomplishes a spacial transformation of the image, to complete the process, it is necessary to assign intensity values to those new locations of pixels $(x,y)$. For instance, this task is accomplished using \textbf{intensity interpolation}. Two of the often used interpolation methods are Nearest Neighbor Interpolation and Bilinear Interpolation. Each one has its computational cost. For Bilinear Interpolation, the assigned value is obtained using linear interpolation first in one direction, and then again in the other direction, or using the Equation~\eqref{bilinear-interpolation}
\begin{equation}\label{bilinear-interpolation}
v(x,y)=ax+by+cxy+d
\end{equation}
where the four coefficients $(a,b,c,d)$ are determined by substituting the four nearest neighbors $(x_1,y_1), (x_2,y_2), (x_3,y_3)$ and $ (x_4,y_4)$ of point $(x,y)$ in the Equation~\eqref{bilinear-interpolation}, and solving the linear equation system. 

In general, for a $M\times N$ image $f(x,y)$, the two-dimensional transform denoted by $T(u,v)$ can be expressed in the general form

\begin{equation}\label{genralform-trasformation}
T(u,v)=\sum_{x=0}^{M-1}\sum_{y=0}^{N-1}f(x,y)r(x,y,u,v)
\end{equation}
where $r(x,y,u,v)$ is called a forward transformation kernel. As before, $x$ and $y$ are spatial variables, while $M$ and $N$ are the row and column dimensions of image f. Variables $u$ and $v$ are called transform variables. The name of the transformation $T(u,v)$ will depend on the kernel $r(x,y,v,u)$, for instance, the Discrete Fourier Transform is given by the kernel
\begin{equation*}
r(x,y,u,v)=\frac{1}{\sqrt{MN}}e^{-2\pi i(\frac{ux}{M}+\frac{vy}{N})}.
\end{equation*}
Any transform has its inverse, for instance, the Inverse Discrete Fourier Transform is given by
\begin{equation}
f(x,y)=\sum_{x=0}^{M-1}\sum_{y=0}^{N-1}T(u,v)r'(x,y,u,v)
\end{equation}
with
\begin{equation*}
r'(x,y,u,v)=\frac{1}{\sqrt{MN}}e^{2\pi i(\frac{ux}{M}+\frac{vy}{N})}.
\end{equation*}

\section{\textit{Filters}}
Image processing is illustrated with the use of an explicit filter. An image can be \textbf{filtered} either in the frequency or in the spatial domain. A filter generally uses more than one pixel from the source image for computing each new pixel value. For example, smoothing an image could be modified to simply replace every pixel with the average of its neighboring pixels. Mathematically, the operation associated with a linear filter is called \textbf{linear convolution} and in general combines two functions of the same dimensionality as seen in Equation~\eqref{convolution}.
\begin{equation}\label{convolution}
(w*f)(x,y)=\sum_{s=-a}^{a}\sum_{t=-b}^{b} w(s,t)f(x-s,y-t)
\end{equation}
The function $f(x,y)$ still representing the image and the \textbf{mask function} $w(s,t)$ is a small matrix used for blurring, sharpening, embossing, edge and corner detection, and more. For example, consider the image given by the matrix in Table~\ref{image}, this image will be modified by the mask in Table~\ref{mask} to obtain a new image, the value 4.5 is an approximation value given by the average of the neighbors, that is, by applying the convolution formula to the $3\times 3$ submatrix in the left top of the original image.

\begin{table}[h]
\small
\begin{minipage}{0.3\textwidth}
\centering
    \begin{tabular}{|m{0.3cm}|m{0.3cm}|m{0.3cm}|m{0.3cm}|m{0.3cm}|}
            \hline
            3&1&7&2&1\\ [0.5ex]
            \hline
            2&1&7&2&4\\
            \hline
            5&9&6&6&7\\
            \hline
            2&1&5&7&1\\
            \hline
            3&4&1&7&3\\
            \hline
        \end{tabular}
        \caption{Source  image.}\label{image}
\end{minipage}\hfill 
\begin{minipage}{0.3\textwidth}
\centering
    \begin{tabular}{|c|c|c|}
            \hline
            $\frac{1}{9}$&$\frac{1}{9}$&$\frac{1}{9}$\\
            \hline
            $\frac{1}{9}$&$\frac{1}{9}$&$\frac{1}{9}$\\
            \hline
            $\frac{1}{9}$&$\frac{1}{9}$&$\frac{1}{9}$\\
            \hline
        \end{tabular}
        \caption{Mask.}
        \label{mask}
\end{minipage}
\begin{minipage}{0.3\textwidth}
\centering
    \begin{tabular}{|m{0.3cm}|m{0.3cm}|m{0.3cm}|m{0.3cm}|m{0.3cm}|}
        \hline
        &&&& \\
        \hline
        &4.5&&& \\
        \hline
        &&&& \\
        \hline
        &&&& \\
        \hline
        &&&& \\
        \hline
        \end{tabular}
        \caption{New image}
\end{minipage}
\end{table}
The importance of linear convolution is based on its simple mathematical properties such as commutativity, linearity, and associativity. Filters of course are not limited by linear functions, there are nonlinear filters; ``nonlinear'' means that the source pixel values are combined by some nonlinear function.
Mathematical transforms provide alternate representations of images as linear combinations of a set of basis functions. Those transforms also help in the interpretation of the features of images, and facilitate computation. A filter takes the information in the spatial domain $(x,y)$ and ``translates” it to the frequency domain $(u,v)$. The Gaussian, mean, derivative, and Hessian of Gaussian ITK filters are examples of linear filters. For instance, the mask in Table~\ref{mask} is used for the mean filter (the mean of the neighbors).

\begin{figure}[H]
     \centering
     \begin{subfigure}[b]{0.4\textwidth}
         \centering
         \includegraphics[width=\textwidth]{img_c/bajo_original.jpg}
         \caption{Original image}
     \end{subfigure}
     \hfill
     \begin{subfigure}[b]{0.4\textwidth}
         \centering
         \includegraphics[width=\textwidth,height=6.7cm]{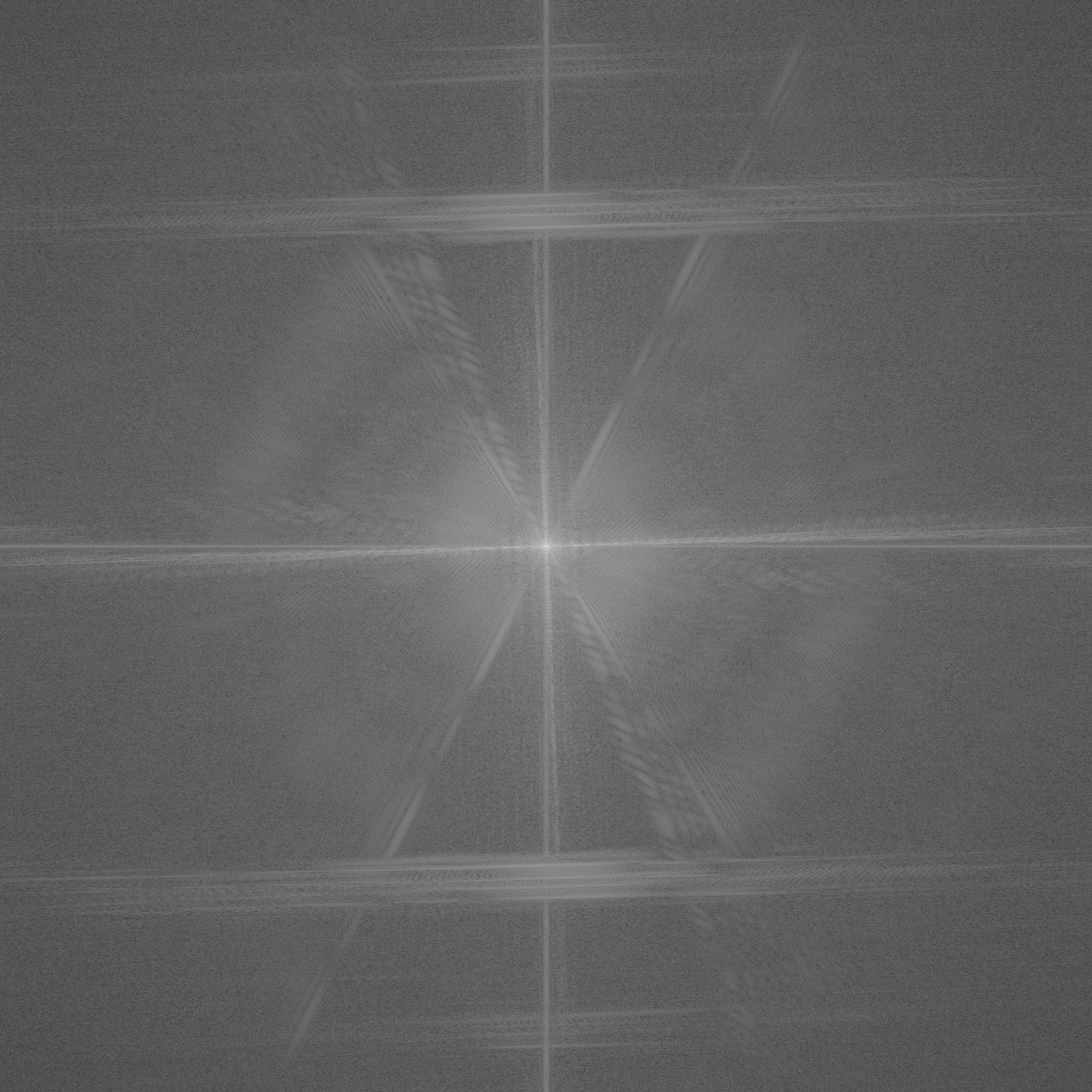}
         \caption{FFT image}
    \end{subfigure}
    \caption{Fast Fourier Trasform FFT}
         \label{fig:FFT}
\end{figure}

Fast algorithms lay out the sequence of computations in such a way that intermediate results are only computed once and optimally reused many times. This ``fast Fourier transform'' or FFT reduces the time complexity of the computation. In the frequency domain, filtering of an image is made by the alteration of its spectrum in a desired way, for example, suppressing the high-frequency components by a low pass filtering (see Figure~\ref{fig:FFT}).

\textbf{Corners} are particularly useful structural elements in video tracking, and stereo images, they are reference points for geometrical measurements, and for calibrating camera systems in machine vision applications. 

There are some corner detectors: Moravec corner detection algorithm, F\"orstner corner detector, multi-scale Harris operator, and others. As an example we describe the Moravec corner detector, it is one of the simplest detectors. Mathematically, the change is characterized in each pixel of the image by $T(u,v)$ which represents the difference between the sub-images for an offset $(u,v)$:
\begin{equation*}
T(u,v)=\sum_{x}\sum_{y}w(u,v)(f(x+u,y+u)-f(x,y))^2
\end{equation*}
where $u$ and $v$ represent the offsets in the four directions
\begin{equation*}
(u,v)\in\{(1,0),(0,1),(1,1),(-1,1)\};
\end{equation*}
the matrix $w(u,v)$ is a rectangular window around the pixel; and $f(x+u,y+u)-f(x,y)$ is the difference between the sub-image $f(x,y)$ and the offset patch $f(x+u,y+u)$. 

In each pixel $(m,n)$, the minimum of $E(u,v)$ in the four directions is kept and denoted by $F_{m,n}$. Finally, the detected corners correspond to the local maxima of $F_{m,n}$, that is, at pixels (m,n) where the smallest value of $E(u,v)$ is large. It is possible to implement the detector using eight offsets to improve the detection.  

It turns out that the Moravec detector has several limitations, then it is not commonly used. Harris corner detector is another more used corner detector, but it is based on the first partial derivatives (gradient) of the image function $f(x, y)$. The algorithm needs to calculate a local structure matrix N, the components of this matrix are convolutioned with a linear Gaussian filter to form a new matrix Ñ. To have relevant parameters for comparison and take the decision if a point is or is not a corner, the eigenvalues of matrix Ñ are calculated. To avoid the explicit calculation of the eigenvalues, the Harris detector defines the function $Q(u, v)$ called the \textbf{corner response function} as a measure of corner strength, which is a weight difference between determinant and trace of matrix Ñ.

\begin{figure}[H]
     \centering
     \begin{subfigure}[b]{0.4\textwidth}
         \centering
         \includegraphics[width=\textwidth]{img_c/bajo_original.jpg}
         \caption{Original image}
     \end{subfigure}
     \hfill
     \begin{subfigure}[b]{0.4\textwidth}
         \centering
         \includegraphics[width=\textwidth,height=6.7cm]{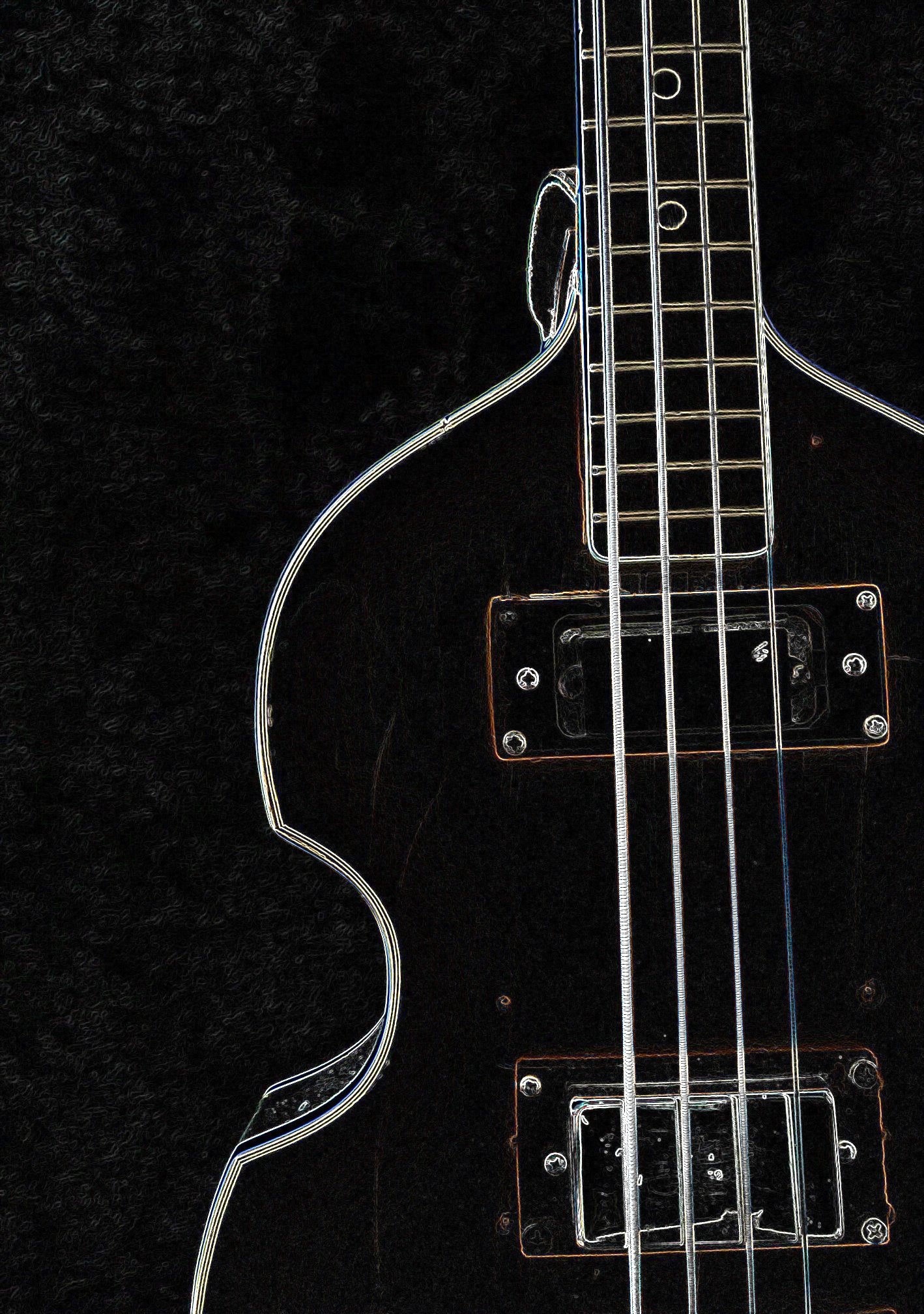}
         \caption{Edges}
    \end{subfigure}
    \caption{Edges enhancing}
         \label{fig:edges}
\end{figure}

Another important topic in image processing is \textbf{detecting edges}. Edge-like structures and contours save computing time in processing of images (see Figure~\ref{fig:edges}). Edges can roughly be described as image positions where the local intensity changes along a particular orientation. The filters for edge detection are based on gradient functions. The applications of this kind of filtering are mainly in computer vision, self-driving cars, and robot guidance.

\section{\textit{Convolutional Neural Networks}}
The image filtering methods early mentioned are essentially mathematical operations of matrices. But nowadays, digital processing goes in another direction. The huge amount of images easily found everywhere can be used as a database { which can be used to train \textbf{Artificial Intelligence} (AI) algorithms for identifying patterns, classification of images, or creating art}. In particular, the neural networks are the basis for the development of new methods to edit images.

This section begin describing how does a Neuronal Network works. 
Consider the problem of binary classification, given input data we want to separate the data into two classes. For example, to separate two kinds of products  in a company, or to verify if the product has the correct weight or not. First assume that the input data are tables of numbers. The aim is to take a vector of numbers to produce a value between 0 and 1. The value is considered to be the probability that the object belongs to a class A. If the value is bigger than 0.5, we select class $A$ as a result; otherwise, we select class $B$. 

\subsection{\textit{Feed Forward Neural Network}}

\textbf{Feed Forward Neural Networks} are used to transform an input vector into a value between 0 and 1. It requires a sequence of steps using two types of transformations: at the $i$ step apply a linear transformation sending the $i$-vector $X_i$ to $Y_i=A_iX_i+B_i$ where $A_i$ is a matrix and $B_i$ is a vector, then apply a non linear transformation such as the $RELU$ which sends the vector $Y_i=(y_1, \cdots, y_n)$ to $X_{i+1}=(RELU(y_1), ...,RELU(y_n))$, where $RELU(y_i)=0$ if $y_i<0$ or $RELU(y_i)=y_i$ otherwise. Note that the last matrix $A_n$ and the last vector $B_n$ determine the dimension of the output, in our case the dimension of the output should be one. The vectors $B_i$ are called \textbf{bias} vectors and they determine the image of the vector $X_i=(0,0,\cdots,0)$.
The important part here is how to find the coefficients of the transformations $A_1,B_1,\cdots, A_n, B_n$ in the feed forward neural network that helps to correctly classify the data. This are called the \textbf{weights} of the algorithm.

The general strategy that will be in use is the following:
\begin{itemize}
\item Collect a data set together with the real class (labels) of each element and split the data set into training data set and test data set. Usually, the test data contains less than 20\% of the total number of data points.
\item Apply the Feed neural network $F$ to a list of data points of the training dataset.
\item Measure how the output $F(p)$, which is a corresponding list of computed classes, differs from the real classes. Call this the loss $L(F)(p)$.
\item \textbf{Back propagation}: Consider the loss $L(F)$ as a function depending on the coefficients of the matrices $A_i$, $B_i$, for all $i$, and the data $p$. Using techniques of calculus, find the direction of maximum increase of the error of $L$ with respect to the weights of $A_n$  and $B_n$ for a fixed $p$ and fixed $A_j$, $B_j$, with $j<n$. That is, find $v_n$ and $v_{n'}$ such that $L(A_n+v_n,B_n+v_{n'})(p)>L(A_n,B_n)(p)$. 
\item Repeat this process for every $i=n-1,\cdots, 1$ to find $v_i$ and $v_{i'}$. In other words, find the direction of maximum increase of the error of $L$ with respect to the weights of $A_i$ and $B_i$ for fixed $p$, $A_j$, $B_j$, with $j\neq i$.
\item The next step is to replace every $A_i$ by $A_i-\alpha v_i$ and $B_i$ by $B_i-\alpha v_{i'}$  reducing the error $L$ of classification of the elements in $p$. Here $\alpha$ is an auxiliary parameter that changes the size of the vectors.
\item Repeat this process until all the data is used.
\item Now measure the error of the algorithm on the test data.
\item According to the results, modify the architecture (change the nonlinear functions, the size of the matrices), increase the data set size, or add regularization based on the plot of the training error and the testing error. See \cite{DLB} for details.
\item Repeat the process until the error on test data and the error on train data are both below acceptable values. 
\end{itemize}

In real life, this process does not have to be done image by image, instead work with batches of images. No algorithms will be trained so this general explanation is enough to follow this chapter content. Curious readers can consult \cite{DLB, Keras, dive} to learn to train their algorithms.

It is important to interpret correctly the error on the training data, and the error on test data. The training error tells if the algorithm is capable of learning the classes of the particular data with which it works. The test error will give an estimation of how well the algorithm will perform in real life, since the test dataset is supposed to be different from the training dataset.

The problem of classification of images is very important, for example, a company would like to distinguish two different products passing in front of a camera. Another possibility is to detect defective products that can be visually distinguished such as corroded or dented cans. A webcam that can detect movement is more useful if it can differentiate between a human and an animal. It is also important in medicine, for example, to classify magnetic resonance imaging with pathologies.  
Consider an image of dimensions $(500,300)$, that is, a $500\times 300\times 3$ matrix. A feed forward neural network would require to transform the matrix into a vector of dimension 5003003.  If the output has dimension one, then, at least it needs to learn the weights of a matrix of size 450000 plus the bias vector. More elaborated architectures require more weights to be determined. The color of a pixel is determined by the position of the pixel and, in the case of a photograph, most pixels share similar colors with nearby pixels. Due to this reason, working with vectors has the inconvenience that geometrical information is lost. When the matrix becomes a vector, it is no longer known what vectors were above/below a particular pixel.

\subsection{Convolutional Neural Network} 
A \textbf{Convolutional Neural Network (CNN)} is an architecture of deep learning in which blocks of filters are applied to an image. The user determines the dimension of the blocks and the way the blocks are connected. Then, the general strategy to train the machine learning algorithm is used until the algorithm learns the values of the entries of the filters that best solve the problem. A pooling operation reduces the size of the filter, by replacing squares of weights with the maximum value (or minimum or average). Blocks of convolutions are usually followed by pooling layers. After several iterations, when the output has a smaller size, a feed forward neural network is attached at the end of the algorithm. The parts of the Feed Forward Neural Network are called dense layers or fully connected layers. See the image below. 
\begin{figure}[H]
\centering
\includegraphics[width=\textwidth]{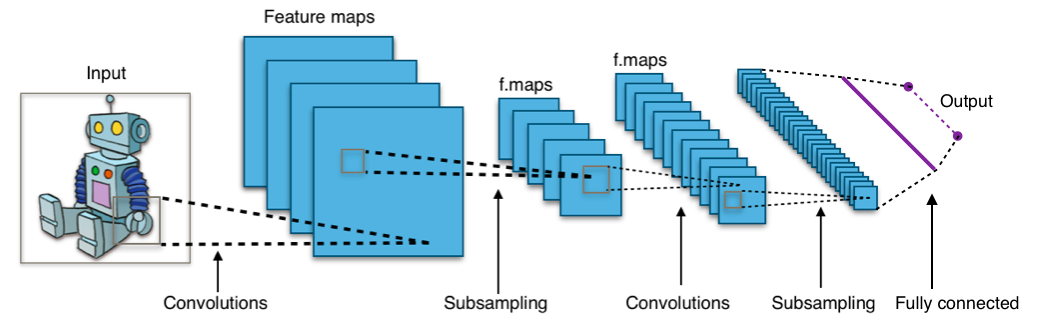}
\caption{Diagram of a standard CNN.}\label{Fig:fromwiki}
\end{figure}

The general strategy to train machine learning algorithms for image data goes as follows: 
\begin{itemize}
\item Collect images with labels
\item Apply the algorithm to the training data set
\item Use a loss function $L$ to measure how far is the output from the real labels of the data.
\item Use back propagation to change the weights of the algorithm.
\end{itemize}
The point is that in the next iteration the classification error should decrease or stay the same. This means that the algorithm evaluated on the training data will give an output similar to the labels of the training data. 

At this point, it is important to measure the success of testing data, that is, the values of the loss function when evaluated on data different from the training data. This is an important step because it will give an idea of how likely it is that the algorithm fails in real life.
Then adjustments are made if needed (add regularization, data augmentation, etc.).
In the end, if the techniques explained in \cite{DLB} are used and with a little luck, the training process will find the weights that produce similar output to the labels on training data and that give a good performance on unseen data.  
For CNN, those weights are the values of the filters. As a consequence, the algorithm find the filters that solve the task. The reader is advised to look at the animation \url{https://adamharley.com/nn_vis/cnn/2d.html} made by Adam Harley. On the top left part, the reader can draw a number, and see how the convolutions and the feed forward algorithm predict a number from 0 to 9.

The reader is not required to know how to program, at the end there are examples of programs where the user is only required to modify certain parameters to create an image. If the reader has experience coding, the language of machine learning is Python. Knowing SQL is also useful to deal with data. The two most important libraries to use are PyTorch and Keras. Both are very easy to use, perhaps PyTorch is more convenient for experiments. Since the reader does not require to know how to code, it is important that the reader familiarizes themself with Google Colab by following the tutorial \url{https://colab.research.google.com/notebooks/welcome.ipynb}.

The reader must try the following two notebooks in which the architectures are introduced. The reader only needs to execute the code. In the first notebook, the reader can submit an image and the network will classify it \url{https://github.com/Rubiel1/Convolutional-Neural-Networks-Applied-to-Modification-of-Images/blob/main/VGG19.ipynb}. In the second notebook \url{https://github.com/Rubiel1/Convolutional-Neural-Networks-Applied-to-Modification-of-Images/blob/main/Lenet.ipynb}
the reader can run a digit classifier.

At the moment of training, to freeze certain weights means that those weights are not updated during the back propagation stage.

\section{Examples of CNN}
The following subsections illustrate how variations on the general strategy to train CNN leads to unforeseen applications.

\subsubsection{CNN for classifier} 
Training a CNN requires some advanced techniques, but in simpler cases, it can be done with auto-machine learning. Here are some popular services:
\begin{itemize}
\item From Uber \url{https://github.com/ludwig-ai/ludwig}.
\item From Keras \url{https://github.com/keras-team/autokeras}.
\item Paid service by Google \url{https://cloud.google.com/automl/} 
\item Paid service by Microsoft  \url{https://docs.microsoft.com/en-us/azure/machine-learning/concept-automated-ml}. 
\end{itemize}


A simple classifier between two objects will be trained in order to settle ideas. The reader should visit \url{https://teachablemachine.withgoogle.com/} and click on get started. Then, the reader should click on image project and then on standard image model.  In the next menu, the reader can either upload a folder of images (at least 300) of each class, or use the webcam to create a dataset. 

\begin{figure}[H]
\centering
\includegraphics[width=\textwidth]{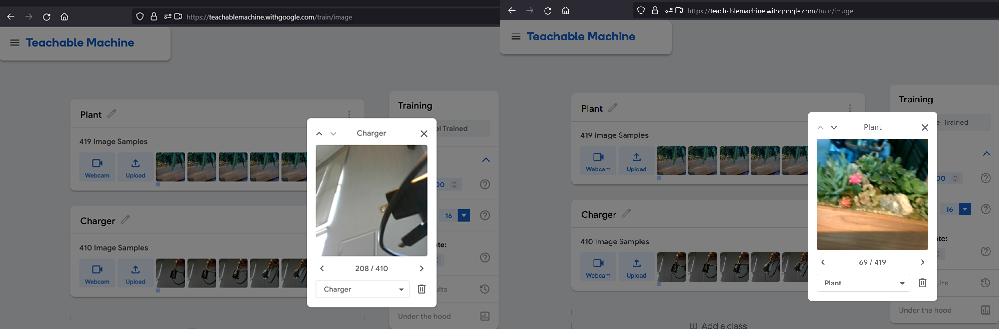}
\caption{Training on Teachable Machine}\label{Fig:training}
\end{figure}

The reader should change the name of the first class to ``Plant'' and use the camera to create a data set of more than 300 images of plants. Then, the reader should change the name of the second class to ``Charger'' and used the camera to create a dataset of more than 300 images of phone chargers. See Figure~\ref{Fig:training}.

After pressing train, and after a couple of minutes, the algorithm is ready to test. On the right side, it is possible to choose between file or camera input.

\begin{figure}[H]
\centering
\includegraphics[width=\textwidth]{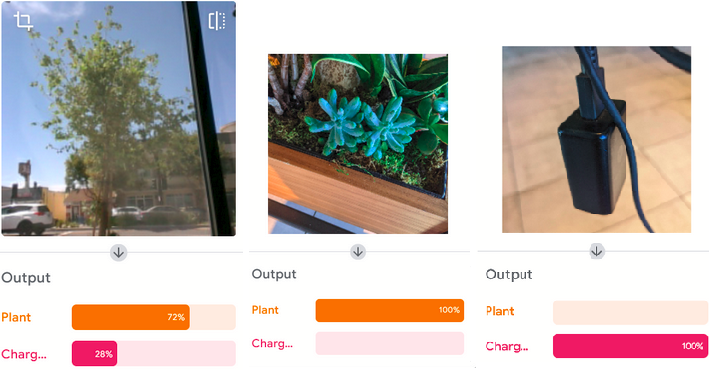}
\caption{Teachable Machine results}\label{Fig:testing}
\end{figure}

The output is understood as the probability that the input image belongs to a certain class. For example, in Figure~\ref{Fig:testing}  the image of the plant and the image of the charger were correctly classified, but the image of the tree has $72\%$ probability of being a plant, while $28\%$ of being a charger. Note that it is possible to put an image of any object and the algorithm will continue to return the probability that the image is a plant versus the probability that the image is a charger. The model used in this book, already trained, is available at \url{https://teachablemachine.withgoogle.com/models/G1tYUnYIT/}. 
Now that a general idea about how convolutions in machine learning allow working with image data is realized, variations on the general strategy can be made to train a machine learning algorithm.

\subsubsection{Adversarial attack.}
An often overlooked aspect of designing and training models of CNN is security and robustness, especially in the face of an adversary who wishes to fool the model.

Consider a pre-trained model, an algorithm that classifies images of numbers. The input is an image and the output is a class: eight, seven, one, etc. Here there is a question on the general strategy for image data: After back propagation, what vector is needed to add to the original image to increase the classification error the most?

Consider the following algorithm that takes a pre-trained neural network that classifies images and an input image:
\begin{itemize}
    \item Input an image to the neural network.
    \item Compare the real label (class) of the image with the predicted label.
    \item Freeze the weights of the algorithm. 
    \item Consider the error function as a function depending on the pixels of the input image.
    \item Use back propagation to find the directional vector to add to the image to maximize this error.
    \item Add to the pixels of the original image the vector to maximize the classification error.
\end{itemize}

The result is an image that eventually is misclassified by the particular CNN, even though the image may still look the same to the human eye. The following colab notebook, made by  Nathan Inkawhich, belongs to the tutorials of PyTorch and executes the previous algorithm. In practice, we do not work one image at a time, and there are some coding details that a beginner can ignore. 
\url{https://github.com/Rubiel1/Convolutional-Neural-Networks-Applied-to-Modification-of-Images/blob/main/fgsm_tutorial.ipynb}
Adversarial attacks have been used to create designs that people can put on their faces to avoid face recognition technologies \cite{adv}.

\subsubsection{Deep dreams}

Take a CNN that was trained to classify animals (for example InceptionV3 \cite{IV3}). Pick some layers. The idea of the team of Alexander Mordvintsev \cite{DD} is to take the general strategy and define the loss function to maximize the norm of the activation functions of those fixed layers. So, if the layers detect mammals, images that contain mammals would give a higher value on those layers than images that do not contain mammals.  Apply back propagation while fixing every weight on the CNN, but modifying the pixels of the input image. In this process, after every update of the image, we create an image that gives higher values on the activation functions of the fixed layers. If the layers detect snakeskin, then snakes will appear on the image.

\begin{figure}[H]
\centering
\includegraphics[width=3cm]{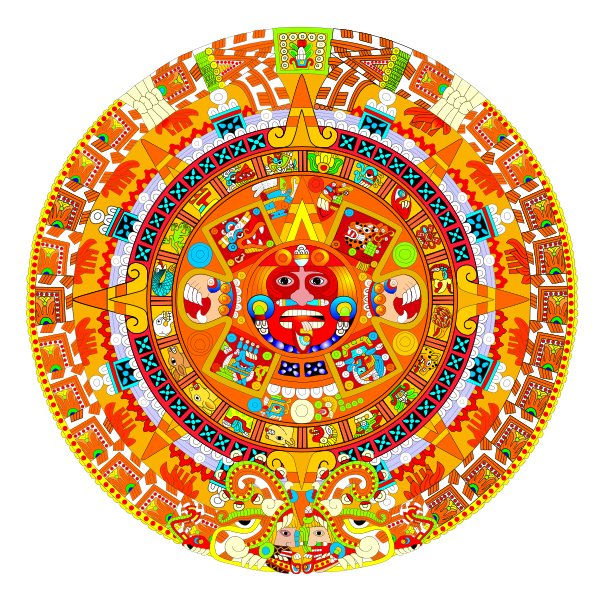}
\caption{Aztec sun stone}\label{Fig:Calendar}
\end{figure}

How to choose which layers to use? To answer that question run an experiment (link at the end of the section) using as an input the Figure~\ref{Fig:Calendar}. Fix Inception V3, and then select one layer at a time to run deep dreams on the calendar image.

\begin{figure}[htb]
     \centering
     \begin{subfigure}[b]{0.2\textwidth}
         \centering
         \includegraphics[width=\textwidth]{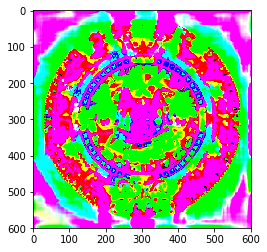}
         \label{fig:area}
     \end{subfigure}
     \hfill
     \begin{subfigure}[b]{0.2\textwidth}
         \centering
         \includegraphics[width=\textwidth]{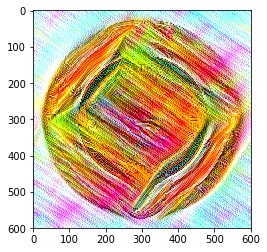}
         \label{fig:square}
     \end{subfigure}
     \hfill
     \begin{subfigure}[b]{0.2\textwidth}
         \centering
         \includegraphics[width=\textwidth]{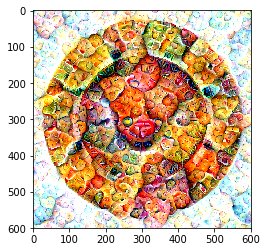}
         \label{fig:groups}
     \end{subfigure}
     \hfill
     \begin{subfigure}[b]{0.2\textwidth}
         \centering
         \includegraphics[width=\textwidth]{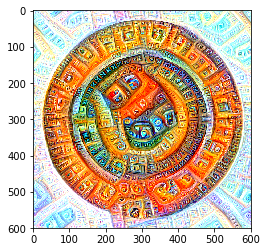}
         \label{fig:regions}
     \end{subfigure}
     \begin{subfigure}[b]{0.2\textwidth}
         \centering
         \includegraphics[width=\textwidth]{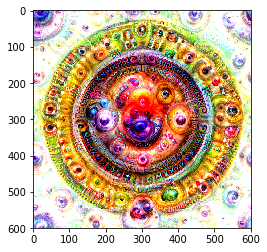}
         \label{fig:eye}
     \end{subfigure}
     \hfill
     \begin{subfigure}[b]{0.2\textwidth}
         \centering
         \includegraphics[width=\textwidth]{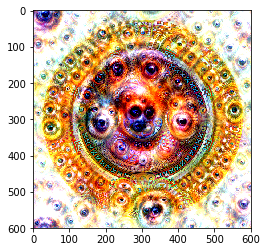}
         \label{fig:eyes}
     \end{subfigure}
     \hfill
     \begin{subfigure}[b]{0.2\textwidth}
         \centering
         \includegraphics[width=\textwidth]{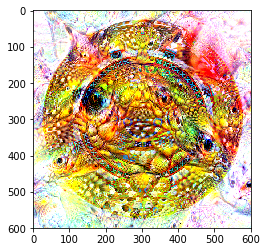}
         \label{fig:snake}
     \end{subfigure}
     \hfill
     \begin{subfigure}[b]{0.2\textwidth}
         \centering
         \includegraphics[width=\textwidth]{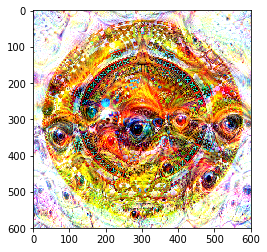}
         \label{fig:human}
     \end{subfigure}
        \caption{Selected output of deep dreams,  fixing one layer at a time of Inception V3, with input the Aztec sun stone.}
        \label{fig:three graphs}
\end{figure}

In  Figure~\ref{fig:three graphs}, the reader can see some examples of the output of deep dreams, going from layers closer to the input to layers closer to the output.

The top left image seems to be separated by regions of colors. If it is compared with the original image, those regions share similar colors on the Aztec sun stone.  As we move to the right images on the first row, the layers learned to detect geometric shapes. To see this note that the white region on the original image now displays lines or even squares. The fifth image contains spheres and so one of the things that the corresponding layer learned is the concept of a sphere. As we move to the right on the second row the sphere becomes animal eyes. Another concept that we can see in the seventh image is the concept of snakeskin.

We conclude that layers closer to the input learn concepts such as colors, curves, lines, and squares. While layers closer to the output learn complex concepts such as eyes, or snakeskin.

The reader can make its experiments by following the instructions in the following jupyter notebook: \url{https://github.com/Rubiel1/Convolutional-Neural-Networks-Applied-to-Modification-of-Images/blob/main/Copy_of_hello_torch_dreams.ipynb} by Mayukh, \url{mayukhdeb.github.io}.
The algorithm is a little bit more complicated, for example, to obtain dreams at different scales it is necessary to resize the images. Details can be found in the notebook.
\subsubsection{Transfer style}

Consider a fixed image from which the user want to learn the style. Perhaps, the image is a photograph of a woodcut or a sketch of a house. Take a second image that provides the content: a landscape, a dancer, a fruit. And a third empty image called the generated image. Take the fixed image, the content image and a generated image, and consider the general strategy where the loss function is replaced by a function that can measure the difference of styles between the generated image and the fixed image, plus the difference in content between the content image and the generated image. Now, freeze everything but the generated pixels of the image. As a consequence, during the training process, modify the generated image pixels to minimize the error, giving it the style of the fixed image, and the content of the second image.  The result is an image that contains the elements of the second image but in the style of the first image.

How are style and content defined? According to the section on deep dreams, the last layers contain complex information about the image. Consider the vector of values of the activation functions on the last layer as the content of the image. To measure how much the content of two images differ, take the norm of the difference of the vectors corresponding to the last layer of each image. Now, style is detected at different scales, and so, pick not only the last layer but several layers of the CNN: $x_{i1},x_{i2},\cdots,x_{in}$. Then consider the Gram matrix whose $jk$ entry is the inner product $<x_{ij},x_{ik}>$. To measure how much the style of two images differ, take the difference between their Gram matrices, then take the norm of that difference. 
The corresponding code is implemented in the following notebook \burl{https://github.com/Rubiel1/Convolutional-Neural-Networks-Applied-to-Modification-of-Images/blob/main/Original_neural_style_tutorial.ipynb}  by Alexis Jacq and Winston Herring. 
\begin{figure}[H]
\centering
\includegraphics[width=\textwidth]{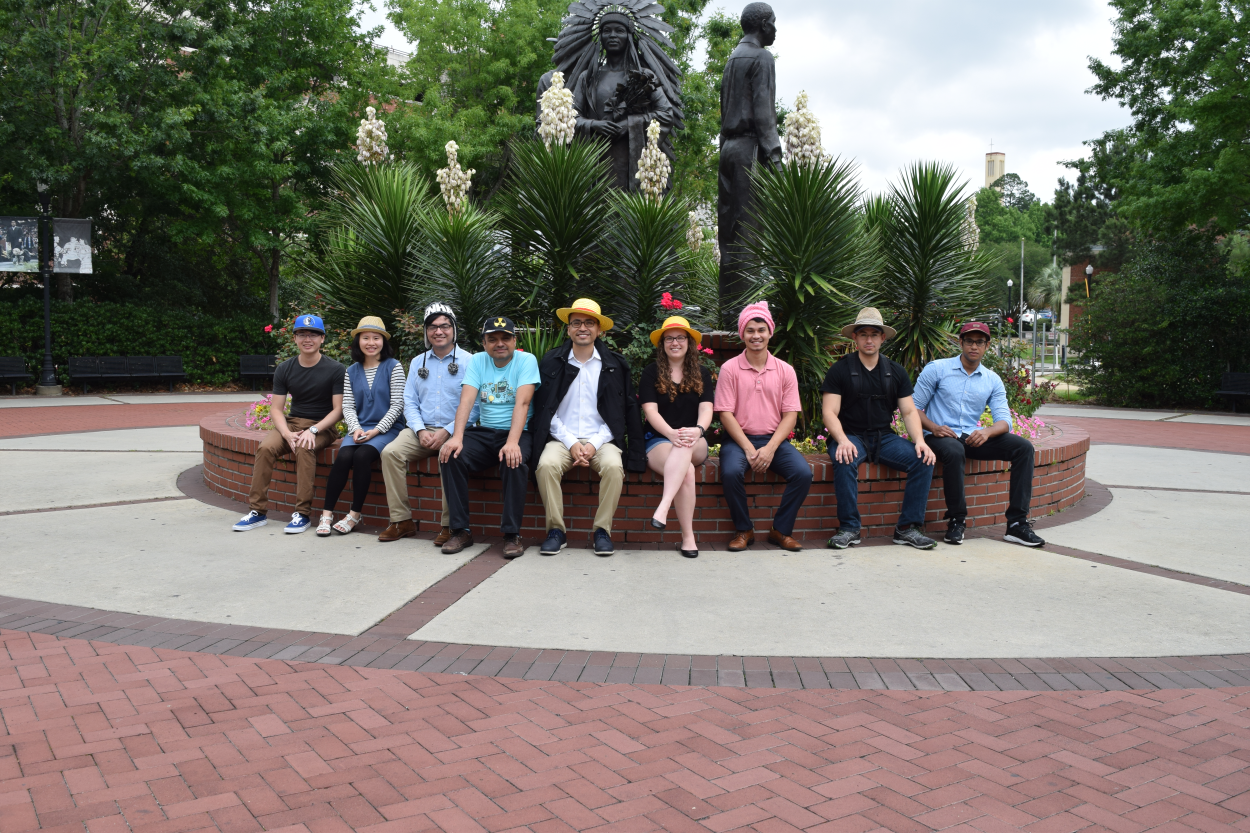}
\caption{The base image is a photograph of Mendoza's Cortes research group}
\end{figure}

\begin{figure}[H]
     \centering
     \begin{subfigure}[b]{0.4\textwidth}
         \centering
         \includegraphics[width=\textwidth]{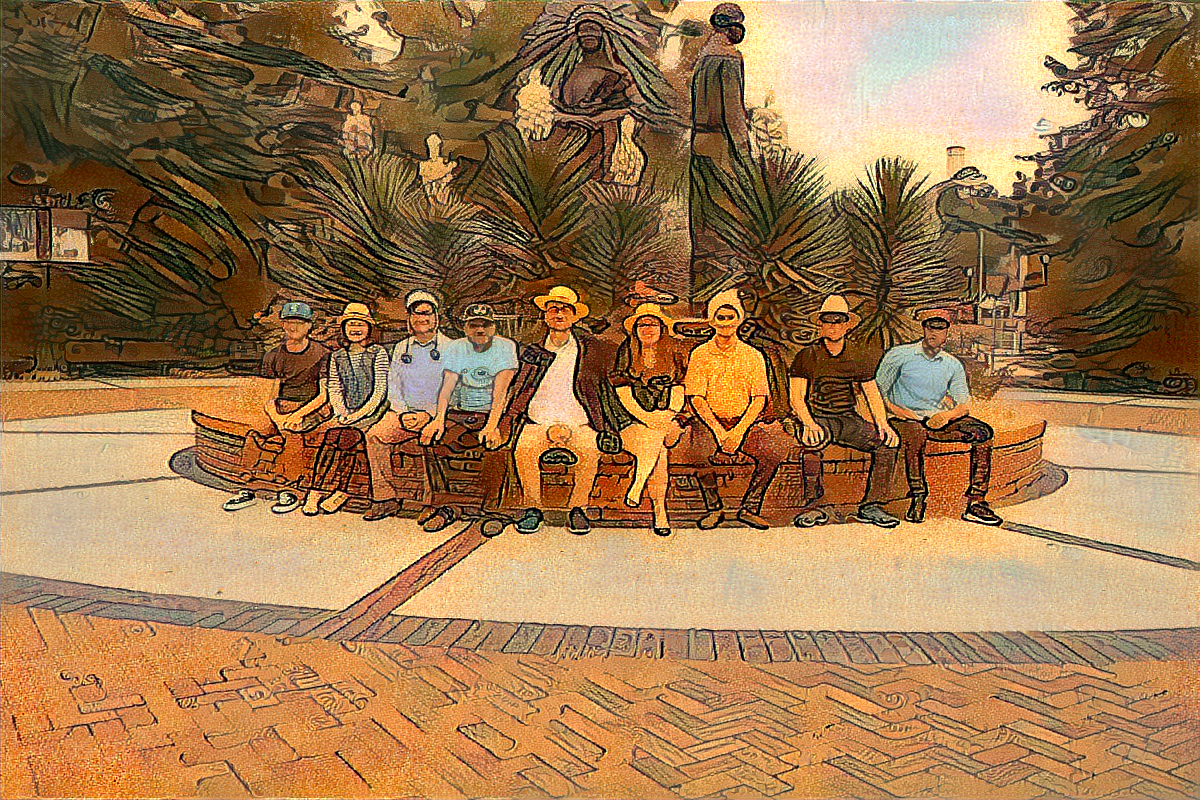}
         \caption{The base image with the style from an Aztec painting}
         \label{fig:Aztec}
     \end{subfigure}
     \hfill
     \begin{subfigure}[b]{0.4\textwidth}
         \centering
         \includegraphics[width=\textwidth]{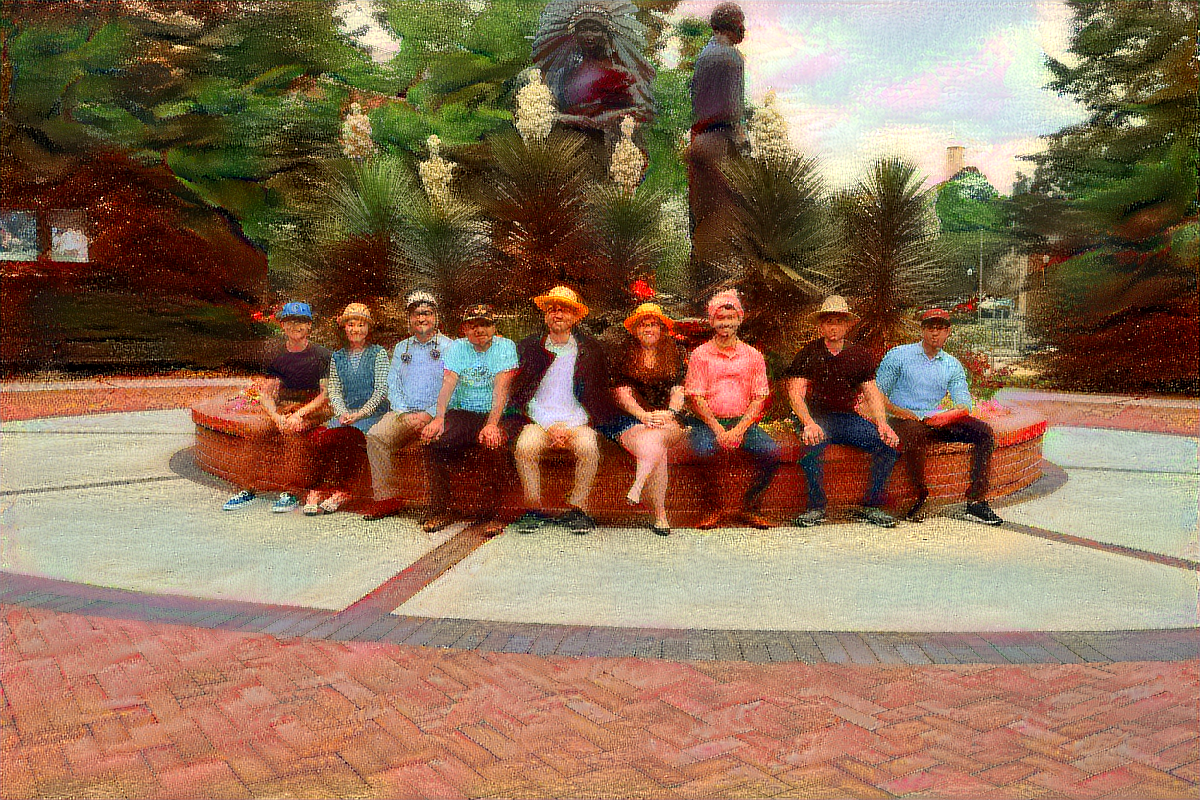}
         \caption{The base image with the style from a Botero painting}
         \label{fig:Botero}
     \end{subfigure}
     \hfill
     \begin{subfigure}[b]{0.4\textwidth}
         \centering
         \includegraphics[width=\textwidth]{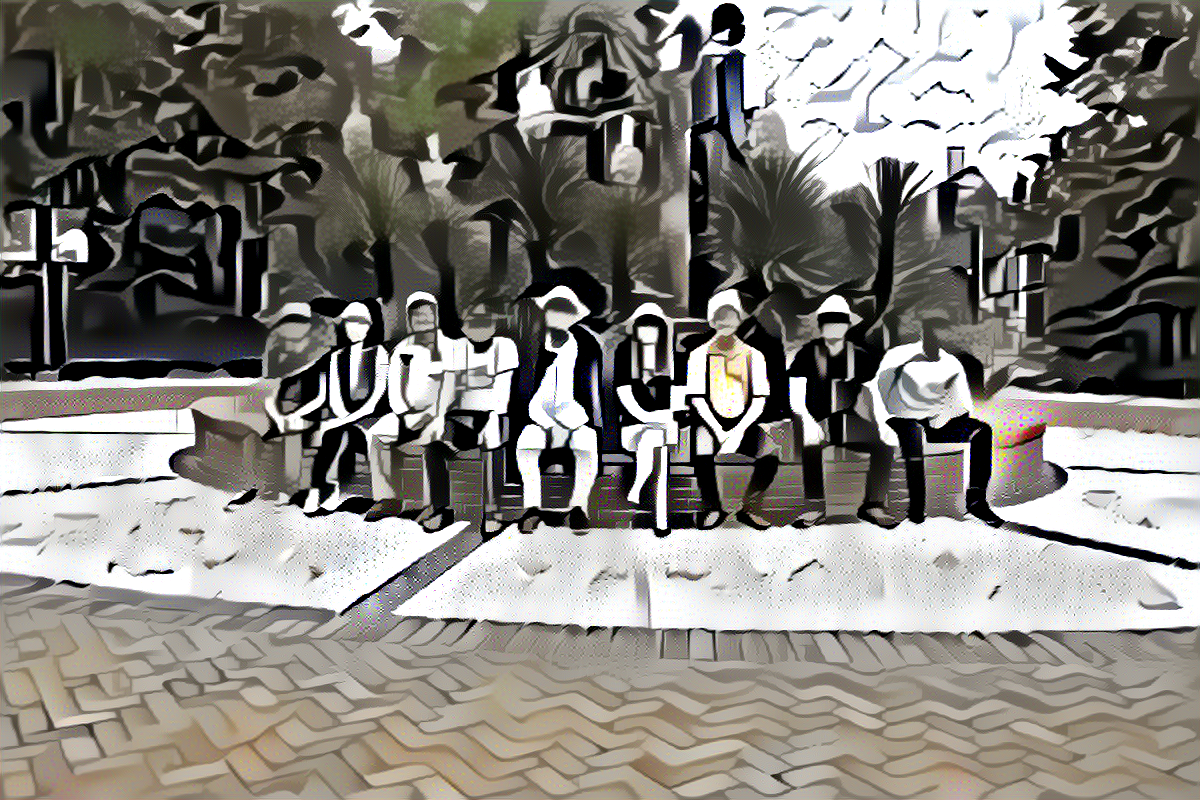}
         \caption{The base image with the style from the Gothic alphabet}
         \label{fig:Gothic}
     \end{subfigure}
     \hfill
     \begin{subfigure}[b]{0.4\textwidth}
         \centering
         \includegraphics[width=\textwidth]{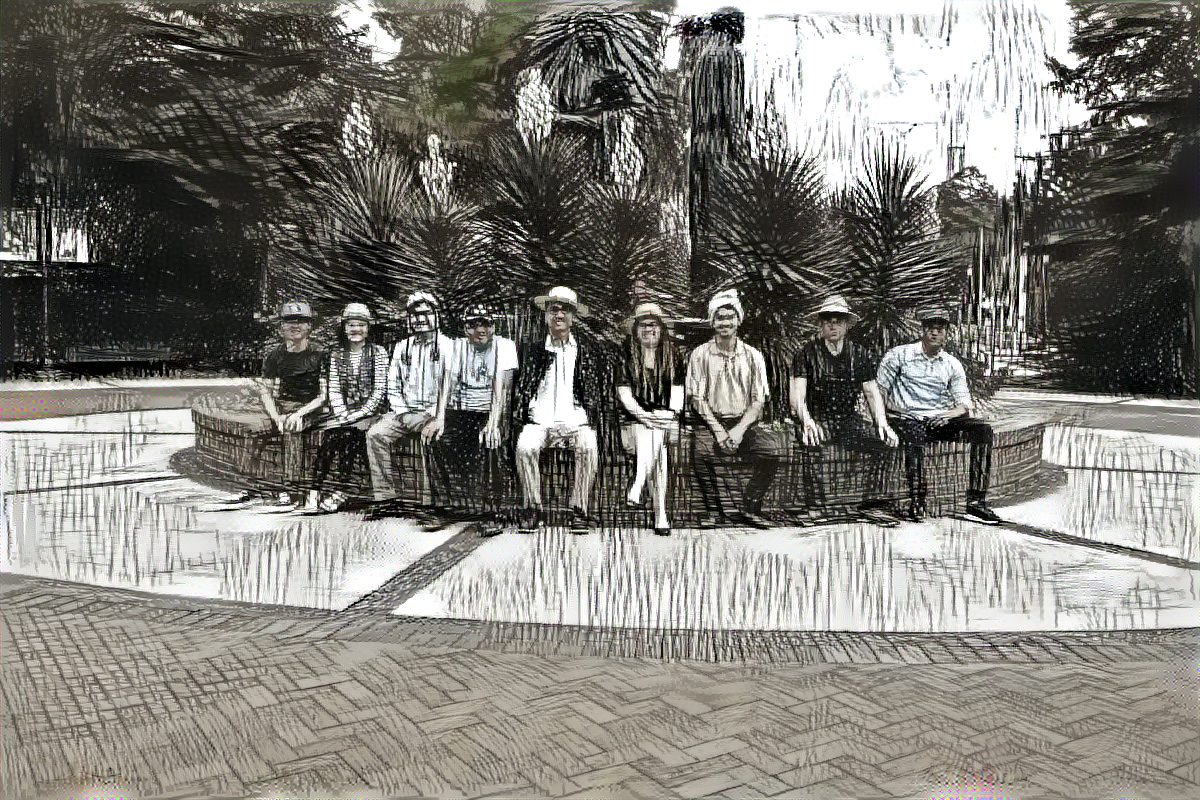}
         \caption{The base image with the style from Woodcut art}
         \label{fig:Woodcut}
     \end{subfigure}
    \hfill
        \caption{Examples of transfer style.}
        \label{fig:Style}
\end{figure}

\subsubsection{Text to Image and diffusion models}
The field of text to image generation aims to create images from a prompt, it was started with the work of \cite{gantti}. People can now write an arbitrary phrase and a machine learning algorithm creates an image related to that text.
 This has immediate applications on the animation of movies and video games as exemplified here~\cite{twittgame}. Prompt engineering refers to the process of learning the key words that make the particular algorithm produce an image with certain characteristics.

The basic idea behind diffusion models is as follows: consider an original image $Image\_0$ and randomly add noise to it, creating $Image\_1$. Then train a neural network to reconstruct  the $Image\_0$ from the damaged $Image\_1$. 

A large group of artist and programmers created the algorithm Disco Diffusion \url{ https://colab.research.google.com/github/alembics/disco-diffusion/blob/main/Disco_Diffusion.ipynb}. Currently, the most advanced diffusion model is Stable Diffusion \cite{dif}. Stable diffusion can be tested at \url{https://stablediffusionweb.com/}. The authors of \cite{dif} first replaced the space of pixels by a lower dimensional space, encoding the images to avoid redundant information, and then, they trained a diffusion model on this new space.
 
\begin{figure}[H]
     \centering
     \begin{subfigure}[b]{0.4\textwidth}
         \centering
         \includegraphics[width=\textwidth]{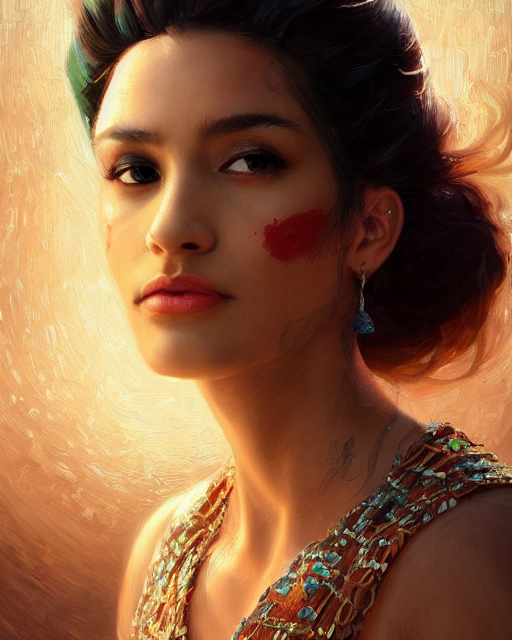}
         \caption{Portrait of latino woman created with Stable diffusion}
         \label{fig:latinoportrait}
     \end{subfigure}
     \hfill
     \begin{subfigure}[b]{0.4\textwidth}
         \centering
         \includegraphics[width=\textwidth,height=5.8cm]{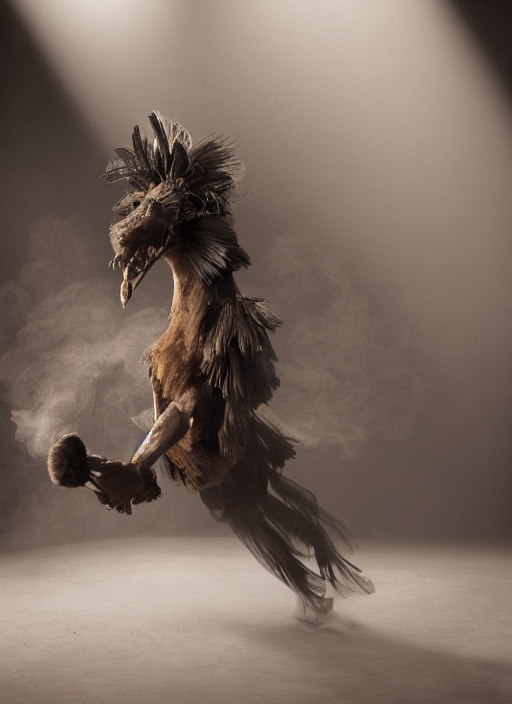}
         \caption{Photograph of a Nahual created with Stable Diffusion}
         \label{fig:Nahual}
     \end{subfigure}
     \caption{}
\end{figure}

AI generated images become popular among the general public with the appearance of Craiyon (initially called DALL-E-mini) in 2022.  Craiyon is an open version, with lower quality, of DALL-E. You can access it here  \url{https://www.craiyon.com/}.
DALL-E and DALL-E2~\cite{dalle} are high quality text to image algorithms. 

Consider the effect of multiplying a vector by a scalar. This operation can change the size of the vector or even make it the zero vector. In machine learning the attention function uses scalar multiplication to determine what parts of the input are more important. DALL-E uses a technology based on the attention mechanism called transformers~\cite{trans}. DALL-E2 added diffusion to obtain better results.

To avoid misuse, DALL-E access is restricted\footnote{https://techcrunch.com/2022/07/20/openai-expands-access-to-dall-e-2-its-powerful-image-generating-ai-system/} and there are key words and outputs that are not allowed. The idea is to prevent targeting minorities or fake nude pictures. Soon, people discovered another reason to keep this technology protected, paper mills can use this technology to create fake images to include in fake research papers, as warned in \cite{twitt}. A new research problem is to create an algorithm that can differentiate real data from fabricated data.
 
Midjourney \url{https://www.midjourney.com} and Stable Diffusion~\url{https://beta.dreamstudio.ai/} are paid services that produce similar quality results to DALL-E2. To prevent their misuse, those algorithms also have filters on the output. New models keep on appearing, Imagen~\cite{google}, Cogview~\cite{cogview2}, to mention the most popular ones. We recommend the dynamic list of resources  \url{ https://pharmapsychotic.com/tools.html} created by the user Pharmapsychotic.

\section{Final remarks}
Here are some important questions that society needs to address.

How will a young artist obtain visibility when this AI tools are already accessible online? 

Are text to image technologies restricting the styles that people will use? 
By using prompt engineering, it is possible to select among the objects that the AI returns. The AI was trained with paintings made by humans, but the new creations are the result of mathematical operations on those human creations. A vector space is defined by fixing a basis and applying addition and multiplication by scalar. The situation in this case is similar, since the paintings  made by the artists are fixed as a basis, and several mathematical operations are applied to combine them into new images. By restricting to AI generated art, the user can only access paintings in this space generated by mathematical operations on the basis. One could argue that when the general public hires an artist, the client also provides instructions and then the artist is the one making the choices to represent those concepts. In that sense, only artist have the full control on how to represent their ideas.

Are these tools perpetuating names and styles that are already popular? 

Consider for example the art of Enrique Brito~\url{http://enriquebritoart.com/about-enrique.html}. Once Enrique Brito creates his paintings, we can learn how to imitate his style. But without examples of paintings by Enrique Brito, no algorithm could create his paintings.
One can argue that some neural networks include probabilistic methods. Then probability allows to expand this theoretical space of generated images using randomness to make a paining with controlled characteristics. However, assuming the creation of paintings with significant differences to what was used to train the algorithms, can such paintings define a new style?

Would the public care that the painting they observe is made by a human or by an algorithm? While this text was being processed, a person has used an algorithm to create a painting, winning a local art competition~\cite{newsai}. 

Who is the author of AI generated art? Artist use internet to promote their work. If an algorithm is trained with their paintings to learn a certain artist's style, new pieces of art can be created without any benefit to the artist. The paintings can even be sold without any benefit to the creators of those algorithms as in \cite{soldai}.

A programmer cannot imagine all the applications of their algorithms, but certainly the struggles of the digital artist were predicted. Which lead to the question, are the programmers doing enough to prevent misuse of this technologies? Since the companies are profiting from the creation of images, they also need to be held accountable for misuses of this technologies.

We end this chapter with a reminder that the purpose of automation is to free humans from repetitive tasks. In \cite{twittgame0} we can see how a rough draft provided by the user, becomes a detailed image. Another benefit is that users that have an idea for a comic, a video game, etc, but do not have the economical means to hire an artist or the talent to draw, those users can now use internet and obtain illustrations for their works as seen in \cite{twittgame}.  While access to internet is still needed, which in some countries is a limit on who can use this technology,  the output of the algorithms has guarantee quality.

\subsubsection{Credits}

The illustrations of Deep Dreams and Style transfer were created by EDC while working on~\cite{fsu}.  We thank Nvidia for the GPU support. EDC was granted access to DALL-E as a user.  EDC was granted access to Stable Diffusion as a researcher. EDC was supported by the National Research Foundation of Korea (NRF) grant funded by the Korea government (MSIT) (No. 2020R1C1C1A01008261). We thank Justus Harris for long discussions about AI generated art.

The image~\ref{Fig:fromwiki} was created by Wikipedia user Aphex34
under the license CC BY-SA 4.0.
The image~\ref{Fig:Calendar} was created by Wikipedia user Manuel Vega Vel\'aquez, CC BY-SA 4.0. 

\bibliographystyle{spbasic}  
\bibliography{author.bib} 

\begin{thebibliography}{21}
\providecommand{\natexlab}[1]{#1}
\providecommand{\url}[1]{{#1}}
\providecommand{\urlprefix}{URL }
\expandafter\ifx\csname urlstyle\endcsname\relax
  \providecommand{\doi}[1]{DOI~\discretionary{}{}{}#1}\else
  \providecommand{\doi}{DOI~\discretionary{}{}{}\begingroup \urlstyle{rm}\Url}\fi
\providecommand{\eprint}[2][]{\url{#2}}

\bibitem[{Bik(2022)}]{twitt}
Bik E (2022) This is the scary stuff {I} have been mentioning in my talks lately. {I} believe fake images like these have already been used by paper mills. we need tools to detect such fakes. \url{https://twitter.com/MicrobiomDigest/status/1536359291266932737}, accessed: 2022-09-03

\bibitem[{Chollet(2021)}]{Keras}
Chollet F (2021) Deep Learning with Python, 2nd edn. Manning Publications

\bibitem[{Ding et~al(2022)Ding, Zheng, Hong, and Tang}]{cogview2}
Ding M, Zheng W, Hong W, Tang J (2022) Cogview2: Faster and better text-to-image generation via hierarchical transformers. arXiv preprint arXiv:220414217

\bibitem[{Eckler(2022{\natexlab{a}})}]{twittgame0}
Eckler D (2022{\natexlab{a}}) Ai art 101. \url{https://twitter.com/daniel_eckler/status/1564601398284664832}, accessed: 2022-09-21

\bibitem[{Eckler(2022{\natexlab{b}})}]{twittgame}
Eckler D (2022{\natexlab{b}}) Stable diffusion is only 30 days old. \url{https://twitter.com/daniel_eckler/status/1572210382944538624}, accessed: 2022-09-21

\bibitem[{Falcon(2018)}]{soldai}
Falcon W (2018) what happens now that an ai-generated painting is sold for \$432,500? Forbes \urlprefix\url{https://www.forbes.com/sites/williamfalcon/2018/10/25/what-happens-now-that-an-ai-generated-painting-sold-for-432500/?}

\bibitem[{Gonzalez and Woods(2008)}]{Digital}
Gonzalez RC, Woods RE (2008) Digital image processing, 3rd edn. Prentice Hall, Upper Saddle River, N. J.

\bibitem[{Goodfellow et~al(2016)Goodfellow, Bengio, and Courville}]{DLB}
Goodfellow I, Bengio Y, Courville A (2016) Deep Learning. MIT Press, \url{http://www.deeplearningbook.org}

\bibitem[{Janesick et~al(1987)Janesick, Elliott, Collins, Blouke, and Freeman}]{scientific}
Janesick JR, Elliott T, Collins S, Blouke MM, Freeman J (1987) {Scientific Charge-Coupled Devices}. Optical Engineering 26(8):268,692, \doi{10.1117/12.7974139}, \urlprefix\url{https://doi.org/10.1117/12.7974139}

\bibitem[{Mendoza-Cortes et~al(2019)Mendoza-Cortes, Dolores-Cuenca, Aguirre, Tzeng, Xu, Jurado, Woerner, Crock, Young, Aduenko, Fuentes, and Chung}]{fsu}
Mendoza-Cortes J, Dolores-Cuenca E, Aguirre C, Tzeng YY, Xu G, Jurado G, Woerner P, Crock N, Young S, Aduenko A, Fuentes MM, Chung H (2019) Machine learning for humans. \url{https://mendozacortesgroup.github.io/MachineLearningForHumans/}, accessed: 2022-09-03

\bibitem[{Mordvintsev et~al(2015)Mordvintsev, Olah, and Tyka}]{DD}
Mordvintsev A, Olah C, Tyka M (2015) Deepdream - a code example for visualizing neural networks. \url{https://web.archive.org/web/20150708233542/http://googleresearch.blogspot.co.uk/2015/07/deepdream-code-example-for-visualizing.html}, accessed: 2022-08-31

\bibitem[{Ramesh et~al(2021)Ramesh, Pavlov, Goh, Gray, Voss, Radford, Chen, and Sutskever}]{dalle}
Ramesh A, Pavlov M, Goh G, Gray S, Voss C, Radford A, Chen M, Sutskever I (2021) Zero-shot text-to-image generation. ArXiv preprint arXiv:2102.12092

\bibitem[{Reed et~al(2016)Reed, Akata, Yan, Logeswaran, Schiele, and Lee}]{gantti}
Reed S, Akata Z, Yan X, Logeswaran L, Schiele B, Lee H (2016) Generative adversarial text to image synthesis. arXiv preprint arXiv:160505396

\bibitem[{Rombach et~al(2022)Rombach, Blattmann, Lorenz, Esser, and Ommer}]{dif}
Rombach R, Blattmann A, Lorenz D, Esser P, Ommer B (2022) High-resolution image synthesis with latent diffusion models. ArXiv preprint arXiv:2112.10752

\bibitem[{Rosenfeld(1969)}]{Picture}
Rosenfeld A (1969) Picture Processing by Computer. Academic Press, New York

\bibitem[{Saharia et~al(2022)Saharia, Chan, Saxena, Li, Whang, Denton, Ghasemipour, Ayan, Mahdavi, Lopes, Salimans, Ho, Fleet, and Norouzi}]{google}
Saharia C, Chan W, Saxena S, Li L, Whang J, Denton E, Ghasemipour SKS, Ayan BK, Mahdavi SS, Lopes RG, Salimans T, Ho J, Fleet DJ, Norouzi M (2022) Photorealistic text-to-image diffusion models with deep language understanding. ArXiv preprint arXiv:2205.11487

\bibitem[{Szegedy et~al(2015)Szegedy, Liu, Jia, Sermanet, Reed, Anguelov, Erhan, Vanhoucke, and Rabinovich}]{IV3}
Szegedy C, Liu W, Jia Y, Sermanet P, Reed S, Anguelov D, Erhan D, Vanhoucke V, Rabinovich A (2015) Going deeper with convolutions. In: 2015 IEEE Conference on Computer Vision and Pattern Recognition (CVPR), pp 1--9, \doi{10.1109/CVPR.2015.7298594}

\bibitem[{Uszkoreit(2017)}]{trans}
Uszkoreit J (2017) Transformer: A novel neural network architecture for language understanding. \url{https://ai.googleblog.com/2017/08/transformer-novel-neural-network.html}, accessed: 2022-08-31

\bibitem[{Vincent(2022)}]{newsai}
Vincent J (2022) An ai-generated artwork’s state fair victory fuels arguments over ‘what art is’. The Verge \urlprefix\url{https://www.theverge.com/2022/9/1/23332684/ai-generated-artwork-wins-state-fair-competition-colorado}

\bibitem[{Yin et~al(2021)Yin, Wang, Yao, Guo, Kong, Ding, Li, and Liu}]{adv}
Yin B, Wang W, Yao T, Guo J, Kong Z, Ding S, Li J, Liu C (2021) Adv-makeup: A new imperceptible and transferable attack on face recognition. ArXiv preprint arXiv:2105.03162

\bibitem[{Zhang et~al(2021)Zhang, Lipton, Li, and Smola}]{dive}
Zhang A, Lipton ZC, Li M, Smola AJ (2021) Dive into deep learning. ArXiv preprint arXiv:2106.11342

\end{thebibliography}

\end{document}